\documentclass[12pt, onecolumn, draftclsnofoot, journal]{IEEEtran}
\usepackage{amsmath,amsfonts}

\usepackage{algorithm}
\usepackage{array}
\usepackage[caption=false,font=normalsize,labelfont=sf,textfont=sf]{subfig}
\usepackage{textcomp}
\usepackage{stfloats}
\usepackage{url}
\usepackage{verbatim}
\usepackage{graphicx} 
\usepackage{cite}
\usepackage{cleveref}
\usepackage{amsmath}     
\usepackage{amssymb}    
\usepackage{booktabs}
\usepackage{amsfonts}
\usepackage{algorithm}
\usepackage{algpseudocode}
\usepackage{tabularx,booktabs}
\usepackage{booktabs}
\usepackage{amsmath}
\usepackage{threeparttable}
\usepackage{xcolor}
\AtBeginDocument{
  \setlength{\abovedisplayskip}{3pt plus 1pt minus 1pt}
  \setlength{\belowdisplayskip}{3pt plus 1pt minus 1pt}
  \setlength{\abovedisplayshortskip}{0pt plus 1pt}
  \setlength{\belowdisplayshortskip}{0pt plus 1pt}
}
\begin{document}

\title{Attention-Based Variational Framework for Joint and Individual Components Learning with Applications in Brain Network Analysis}

\author{Yifei~Zhang,
Meimei~Liu,
and~Zhengwu~Zhang%
\thanks{Y. Zhang is with the Department of Biostatistics,
Yale School of Public Health, New Haven, CT, USA
(e-mail: yifei.zhang.yz2489@yale.edu).}%
\thanks{M. Liu is with the Department of Statistics,
Virginia Tech, Blacksburg, VA, USA
(e-mail: meimeiliu@vt.edu).}%
\thanks{Z. Zhang is with the Department of Statistics and Operations Research,
University of North Carolina at Chapel Hill, Chapel Hill, NC, USA
(e-mail: zhengwu\_zhang@unc.edu). Corresponding author: Zhengwu Zhang.} 
}

\markboth{ArXiv Preprint}%
{Zhang \MakeLowercase{\textit{et al.}}: Attention-Based Variational Framework for Joint and Individual Components Learning}

\maketitle

\begin{abstract}
Brain organization is increasingly characterized through multiple imaging modalities, most notably structural connectivity (SC) and functional connectivity (FC). Integrating these inherently distinct yet complementary data sources is essential for uncovering the cross-modal patterns that drive behavioral phenotypes. However, effective integration is hindered by the high dimensionality and non-linearity of connectome data, complex non-linear SC--FC coupling, and the challenge of disentangling shared information from modality-specific variations. To address these issues, we propose the \textbf{C}ross-\textbf{M}odal \textbf{J}oint-\textbf{I}ndividual \textbf{V}ariational \textbf{Net}work (CM-JIVNet), a unified probabilistic framework designed to learn factorized latent representations from paired SC--FC datasets. Our model utilizes a multi-head attention fusion module to capture non-linear cross-modal dependencies while isolating independent, modality-specific signals. Validated on Human Connectome Project Young Adult (HCP-YA) data, CM-JIVNet demonstrates superior performance in cross-modal reconstruction and behavioral trait prediction. By effectively disentangling joint and individual feature spaces, CM-JIVNet provides a robust, interpretable, and scalable solution for large-scale multimodal brain analysis.
\end{abstract}

\begin{IEEEkeywords}

Brain connectivity, Variational autoencoder, Structural-functional connectivity, Multimodal integration, Joint and individual components.

\end{IEEEkeywords}

\section{Introduction}
\IEEEPARstart{M}{odern} neuroimaging yields complementary views of brain organization through functional connectivity (FC), derived from fMRI, and structural connectivity (SC), via dMRI \cite{Mueller2005}. FC captures temporal correlations in brain activity, while SC delineates the underlying white matter infrastructure. A central goal in neuroscience is to understand the intricate relationship between these modalities to decipher how anatomical structure shapes functional dynamics \cite{Honey2009, Calhoun2016}. While direct anatomical links contribute to functional coupling \cite{Koch2002, Greicius2009}, substantial evidence indicates that complex, network-wide interactions and indirect pathways mediated by SC are crucial for explaining the full repertoire of FC patterns \cite{Suarez2020}.

Significant research efforts have aimed to model this SC-FC relationship using diverse computational strategies, ranging from biophysical simulations to graph-theoretic analyses and statistical learning techniques \cite{Deco2013JNeurosci, Goni2014, Messe2014, Wang2019PLoSCompBio, Popp2024}. Although these methods provide valuable insights, they often face difficulties in fully predicting empirical FC from SC, typically explaining only a moderate fraction of its variance and highlighting limitations in capturing the system's non-linearity and complexity \cite{Ackerman2024}. Concurrently, techniques like Joint and Individual Variation Explained (JIVE) \cite{Lock2013, Murden2022} have focused specifically on disentangling joint and modality-specific sources of variance, yet their effectiveness can be constrained by inherent linear assumptions \cite{Zhou2016}. These challenges, specifically predictive ceilings in modeling SC-FC coupling and linearity constraints in variance decomposition, motivate the development of powerful, flexible frameworks leveraging non-linear latent variable modeling.

Deep generative models, particularly Variational Autoencoders (VAEs) and Graph Autoencoders (GATEs) \cite{Liu2021, Yang2022, Zhang2024MotionInvariant}, represent a promising direction, offering the ability to learn compact, non-linear latent representations of high-dimensional connectome data. However, their application to the \textit{joint} analysis of FC and SC remains underdeveloped. Existing deep learning efforts, while showing progress in specific tasks, often face several drawbacks when considering comprehensive multimodal integration: 
(i) They typically focus on single modality modeling or unidirectional prediction, such as predicting FC from SC. For example, Multilayer Perceptron (MLP)-based approaches have demonstrated strong predictive performance—reporting group-average correlations up to 0.9 and individual-level correlations around 0.55 for FC prediction from SC \cite{Sarwar2021}.  However, this unidirectional framework neglects the simultaneous, bidirectional modeling and joint generation necessary for understanding the interdependent nature of FC–SC relationships.
(ii) The architectures employed, including standard MLPs as used by \cite{Sarwar2021}, are frequently generic and not specifically adapted for the distinct characteristics and structures (e.g., graph topology, symmetry, sparsity) of FC and SC data, potentially limiting representational capacity and biological interpretability. 
(iii) Crucially, they generally lack explicit mechanisms to disentangle joint and modality-specific availabilities, which is vital for a nuanced understanding of multimodal brain organization.

This work addresses two interconnected objectives in multi-modal connectome analysis: (1) to learn interpretable latent representations that factorize paired SC-FC data into joint and individual components while maintaining high-fidelity reconstruction, and (2) to leverage these disentangled representations for enhanced analysis of structure-function relationships and improved prediction of individual cognitive traits.

More specifically, we propose the \textbf{C}ross-\textbf{M}odal \textbf{J}oint-\textbf{I}ndividual \textbf{V}ariational \textbf{Net}work (CM-JIVNet). This novel framework is designed for simultaneous joint generation and analysis of FC and SC, integrating connectivity-tailored deep learning architectures with an explicit attention-based mechanism for latent space disentanglement. Our key contributions are:

\begin{enumerate}
    \item A joint generation and disentanglement framework: CM-JIVNet is a unified VAE architecture capable of generating both FCs and SCs. Crucially, it can disentangle these networks' latent representations into components reflecting \textit{joint} cross-modal information and \textit{modality-specific} variations.
    \item Connectivity-tailored architectures: We employ specialized encoder and decoder structures (leveraging Convolutional Neural Networks (CNNs) and Residual Networks (ResNets), adapted for graph-structured data) optimized for the distinct structures of FC and SC data, enhancing representational power and generative fidelity.
    \item Attention-based latent space fusion and separation: We incorporate a Query-Key-Value (QKV) multi-head attention mechanism directly within the latent space. This module dynamically fuses information to form the joint latent representation while also isolating the independent latent variables for each modality, enabling effective modeling of joint variance and specific features.
    \item Comprehensive empirical validation: We rigorously evaluate CM-JIVNet on a large cohort of 1,065 subjects from the Human Connectome Project Young Adult (HCP-YA) dataset. We demonstrate state-of-the-art performance through significantly lower Fréchet Inception Distance (FID) scores for generative quality, improved accuracy in cross-modality prediction, and behavioral traits prediction compared to relevant baselines.
\end{enumerate}

\section{Datasets and Preprocessing}\label{sec:data}

\subsection{HCP Dataset}
The Human Connectome Project Young Adult (HCP-YA) \cite{VanEssen2013} is a large‐scale effort that provides high‐quality T1-weighted MRI,  dMRI, and resting‐state fMRI (rs-fMRI) data, along with behavioral assessments, for over 1,200 healthy adults. In this study, we processed 1,065 subjects (age range: 22–35 years) who had both T1 and dMRI available. Below, we briefly outline our image preprocessing pipeline for deriving FC and SC matrices.
\subsection{Data Preprocessing}
\subsubsection{Structural Connectivity}
We processed minimally preprocessed dMRI and T1 data for each HCP-YA subject using the population‐based structural connectome (PSC) mapping framework \cite{Zhang2018}. The dMRI acquisition comprised six runs with three diffusion‐weighting shells (b = 1000, 2000, and 3000 s/mm\textsuperscript{2}), each collected with opposing phase‐encoding polarities. This yielded approximately 90 gradient directions per shell and six interspersed b\textsubscript{0} volumes per run (270 diffusion‐weighted volumes total) at 1.25 mm isotropic resolution \cite{VanEssen2012}. T1-weighted images were acquired at 0.7 mm isotropic resolution \cite{VanEssen2012}. 
PSC leverages a reproducible probabilistic tractography algorithm \cite{MaierHein2017} constrained by anatomical priors from the T1 image to reduce bias in tract reconstruction. We parcellated the cortex into 68 regions of interest (ROIs) using the Desikan–Killiany atlas \cite{Desikan2006}. To extract streamlines, gray‐matter ROIs were dilated into adjacent white matter, streamlines traversing multiple ROIs were segmented to isolate complete pathways, and outlier tracts were removed based on length and curvature criteria. Finally, structural connectivity strength between each ROI pair was quantified by the number of reconstructed streamlines, a measure widely used in imaging–genetics studies \cite{Zhao2022}.

\subsubsection{Functional Connectivity}
The HCP-YA rs-fMRI data include two left-right and two right-left phase‐encoded, eyes‐open runs of 15 min each \cite{VanEssen2012}. Imaging parameters were 2 mm\textsuperscript{3} isotropic voxels with 0.72 s repetition time (TR). For each run, we extracted the average BOLD time series for each of the 68 cortical ROIs defined by Desikan et al \cite{Desikan2006}. Pairwise Pearson correlation coefficients were computed between all ROI time courses, Fisher \(z\)-transformed, averaged across the four runs, and then inverse-transformed back to correlation values.

\subsection{Data Representation}
We represent brain connectivity as networks where nodes correspond to anatomically defined ROIs and edges encode connectivity strength between region pairs. For each subject $i$, structural connectivity is denoted as a symmetric matrix $X_{\text{SC}}^{i} \in \mathbb{R}^{V \times V}$, where entry $X_{\text{SC}}^{i}[uv]$ quantifies the number of white matter fiber tracts connecting regions $u$ and $v$. The symmetry property $X_{\text{SC}}^{i}[uv] = X_{\text{SC}}^{i}[vu]$ reflects the bidirectional nature of anatomical connections. Functional connectivity is similarly represented as a symmetric matrix $X_{\text{FC}}^{i} \in \mathbb{R}^{V \times V}$, where each entry $X_{\text{FC}}^{i}[uv]$ denotes the temporal correlation (typically Pearson correlation) between BOLD signal time series from regions $u$ and $v$, satisfying the symmetric property that $X_{\text{FC}}^{i}[uv] = X_{\text{FC}}^{i}[vu]$. We further vectorize each symmetric matrix by extracting its lower triangular elements, excluding the diagonal. An analogous vectorization is also applied to functional connectivity. With a bit of notation abuse, we also use $X_{\text{SC}}^{i}$ and $X_{\text{FC}}^{i}$ to denote the vectorized SC and FC for the subject $i$, respectively.

\section{Method}
Given a dataset $\mathcal{D} = \{(X_{\text{SC}}^{i}, X_{\text{FC}}^{i})\}_{i=1}^N$ containing paired structural and functional brain connectivity matrices from $N$ subjects, we propose CM-JIVNet to learn interpretable latent representations that capture both modality-joint and modality-specific patterns of brain connectivity. CM-JIVNet addresses two interconnected goals:
(i)  to learn low-dimensional latent factors that jointly encode joint and individual variations across SC and FC while preserving the ability to accurately reconstruct 
each modality; (ii) to leverage these disentangled representations for interpretable characterization of SC–FC coupling and improved prediction of individual traits through multimodal information integration.

\subsection{Model Overview}
\begin{figure*}[ht]
  \centering
  \includegraphics[width=0.8\textwidth]{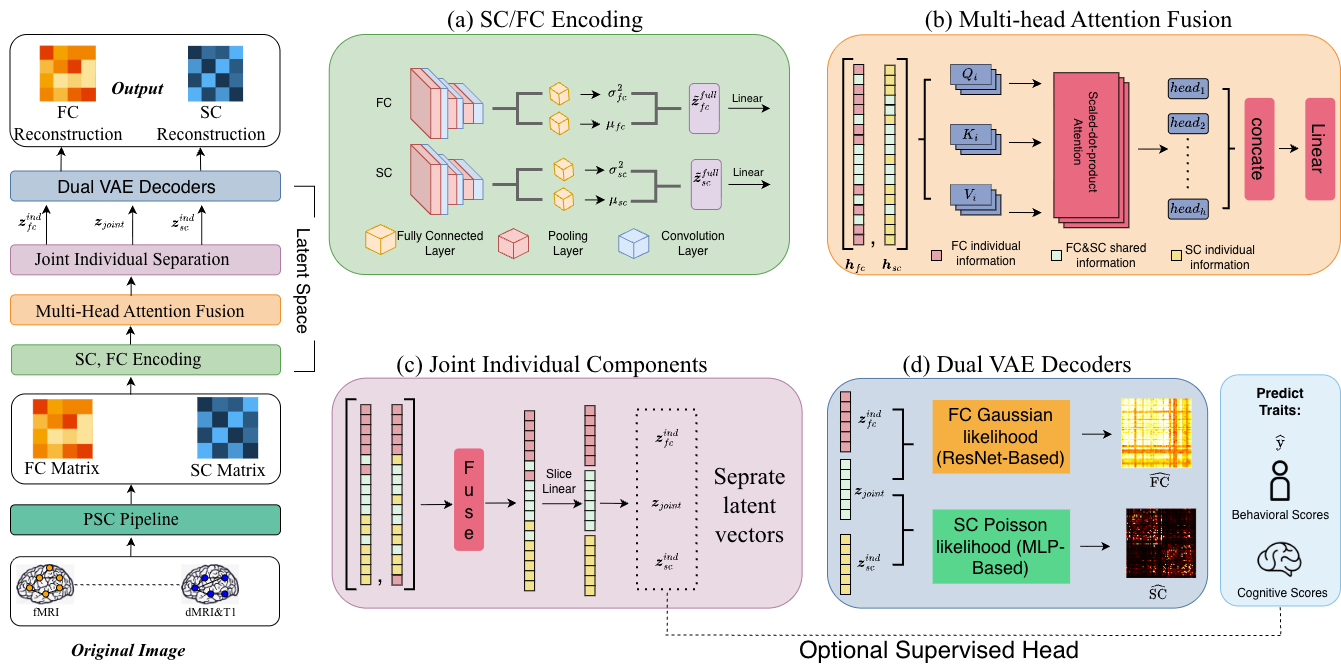}
  \caption{Overview of the CM-JIVNet workflow.
CM-JIVNet comprises four main modules.
(1) {\bf Dual variational encoders} extract latent Gaussian parameters from the FC and SC connectivity matrices.
(2) {\bf Attention-based fusion} aligns the modality-specific latent spaces through a multi-head self-attention and gating mechanism, capturing nonlinear SC–FC dependencies.
(3) {\bf Joint–individual separation} decomposes the fused latent representation into three orthogonal components—joint, FC-specific, and SC-specific—enhancing interpretability of cross-modal coupling.
(4) {\bf Dual decoders and supervision (optional)} reconstruct each modality from its corresponding latent factors while separate branch (optional) encourages prediction of traits from the learned latent factors.
During training, reconstruction and prediction losses are optimized jointly under variational and orthogonality regularization constraints.}
  \label{fig:supervised}
\end{figure*}

An overview of the CM-JIVNet architecture is presented in Fig.~\ref{fig:supervised}.
The model adopts a dual-branch variational autoencoder structure that processes SC and FC connectomes through four sequential modules:

\begin{enumerate}
\item {\bf Dual VAE Encoders}:
Two modality-specific convolutional variational encoders independently process the FC and SC matrices to extract probabilistic latent representations.
Each encoder outputs Gaussian distribution parameters $(\boldsymbol{\mu}_{\text{fc}}, \boldsymbol{\sigma}_{\text{fc}}^2)$ and $(\boldsymbol{\mu}_{\text{sc}}, \boldsymbol{\sigma}_{\text{sc}}^2)$, enabling stochastic latent sampling via the reparameterization trick.

\item {\bf Attention-Based Cross-Modal Fusion}:
A multi-head self-attention and gating fusion module integrates the two latent spaces, identifying nonlinear dependencies between SC and FC.

\item {\bf Joint–Individual Separation Module}:
The fused latent representation is decomposed into three statistically orthogonal components, i.e., joint, FC-specific, and SC-specific components, through an attention-guided separation mechanism.
This decomposition extends the classical JIVE framework to a nonlinear variational setting, promoting interpretability by disentangling modality-common from modality-unique patterns.

\item {\bf Dual VAE Decoder and Supervision}:
CM-JIVNet employs two modality-specific VAE decoders to reconstruct functional and structural connectivity data using likelihood models tailored to each modality. The FC decoder is implemented with a Gaussian likelihood and a ResNet-based architecture, whereas the SC decoder uses a Poisson likelihood parameterized by an MLP-based network. These decoders jointly enforce modality-consistent reconstruction and serve as generative constraints for the latent representations. To introduce a trait-supervised version of CM-JIVNet, a two-stage training procedure is applied. The model is first trained using only reconstruction and regularization losses. A supervised trait-prediction head is then added and fine-tuned, during which both decoders remain active to preserve cross-modal reconstruction consistency.

\end{enumerate}

Together, these four modules form a unified generative framework that disentangles joint and modality-specific representations of brain connectivity, enabling both interpretable cross-modal analysis and predictive modeling.

\subsection{Dual VAE Encoders}\label{sec:VAE}

We model the joint distribution of structural and functional connectivity through a latent variable framework. Specifically, denote $\boldsymbol{z} \in \mathbb{R}^d$ as the latent representation of the paired connectome $(X_{\text{SC}}, X_{\text{FC}})$. For notational simplicity, we drop the subject index $i$; all derivations apply to each subject independently. The generative process is characterized by:
\begin{equation*}
p_{\theta}(X_{\text{SC}}, X_{\text{FC}}) = \int p_{\theta}(X_{\text{SC}}, X_{\text{FC}}|\boldsymbol{z})p(\boldsymbol{z})d\boldsymbol{z}
\end{equation*}
where $p(\boldsymbol{z}) = \mathcal{N}(\mathbf{0}, \mathbf{I})$ represents the standard Gaussian prior over latent variables, and $p_{\theta}(X_{\text{SC}}, X_{\text{FC}}|\boldsymbol{z})$ characterizes how latent representations generate observable connectivity patterns. We assume conditional independence of modalities given $\boldsymbol{z}$:
\begin{equation*}\label{eq:gen:2}
p_{\theta}(X_{\text{SC}}, X_{\text{FC}}|\boldsymbol{z}) = p_{\theta}(X_{\text{SC}}|\boldsymbol{z}) \cdot p_{\theta}(X_{\text{FC}}|\boldsymbol{z}),
\end{equation*}
where the latent variable $\boldsymbol{z}$ captures the coupling between SC and FC while allowing each modality to follow its own generative process.

However, a single entangled latent representation obscures interpretability. Understanding brain connectivity mechanisms requires distinguishing which latent features are shared across modalities and which are modality-specific. Prior work \cite{Honey2009} suggests that SC and FC reflect both overlapping and distinct neural mechanisms: joint components encode fundamental organizational principles constrained by anatomy, while modality-specific components capture features unique to structural wiring or dynamic neural activity. 

We therefore decompose the latent representation $\boldsymbol{z}$ into three interpretable, factorized components: (i) a joint factor $\boldsymbol{z}_{\text{joint}}$ capturing connectivity patterns shared across both modalities, (ii) an SC-specific factor $\boldsymbol{z}_{{\text{sc}}}^{\text{ind}}$ representing anatomical features unique to structure, and (iii) an FC-specific factor $\boldsymbol{z}_{{\text{fc}}}^{\text{ind}}$ encoding functional dynamics independent of structural constraints. The complete latent vector is defined as:
$
\boldsymbol{z} = [\boldsymbol{z}_{\text{joint}}; \boldsymbol{z}_{\text{sc}}^{\text{ind}}; \boldsymbol{z}_{{\text{fc}}}^{\text{ind}}].
$
To achieve JIVE-like disentanglement \cite{Murden2022}, we ensure that each component captures non-overlapping information by enforcing mutual orthogonality between the latent subspaces. Since each latent component is modeled as a Gaussian distribution with mean $\boldsymbol{\mu}$ and variance $\boldsymbol{\sigma}^2$ (detailed in the encoder specification below), we encourage the directions of the posterior mean vectors to be orthonormal:
\begin{equation*}
\langle \hat{\boldsymbol{\mu}}_{\text{a}}, \hat{\boldsymbol{\mu}}_{\text{b}} \rangle = \delta_{\text{ab}}, \quad a,b \in \{\text{joint, sc, fc}\},
\end{equation*}
where $\hat{\boldsymbol{\mu}} = \boldsymbol{\mu} / \|\boldsymbol{\mu}\|$ denotes the normalized mean vector and $\delta_{\text{ab}}$ is the Kronecker delta (0 if $a,b$ are different and 1 if they are the same.)

With this decomposition, we reformulate the generative model by defining modality-specific latent combinations: $\boldsymbol{z}_{\text{sc}}^{\text{full}} = [\boldsymbol{z}_{\text{joint}}; \boldsymbol{z}_{\text{sc}}^{\text{ind}}]$ and $\boldsymbol{z}_{\text{fc}}^{\text{full}} = [\boldsymbol{z}_{\text{joint}}; \boldsymbol{z}_{\text{fc}}^{\text{ind}}]$. This yields:
\begin{equation*}\label{eq:gen:3}
p_{\theta}(X_{\text{SC}}, X_{\text{FC}}|\boldsymbol{z}) = p_{\theta}(X_{\text{SC}}|\boldsymbol{z}_{\text{sc}}^{\text{full}}) \cdot p_{\theta}(X_{\text{FC}}|\boldsymbol{z}_{\text{fc}}^{\text{full}}).
\end{equation*}
Each modality is generated from a combination of joint factors capturing cross-modal dependencies and modality-specific factors capturing unique characteristics.

To learn this factorized generative model from observed data, we employ a VAE framework
that approximates the intractable posterior $p(\boldsymbol{z}_{\text{joint}}, \boldsymbol{z}_{\text{sc}}^{\text{ind}}, \boldsymbol{z}_{\text{fc}}^{\text{ind}}|X_{\text{SC}}, X_{\text{FC}})$ with a Gaussian:
\begin{equation*}\label{eq:posterior}
q_{\phi}(\boldsymbol{z}_{\text{joint}}, \boldsymbol{z}_{\text{sc}}^{\text{ind}}, \boldsymbol{z}_{\text{fc}}^{\text{ind}}|X_{\text{SC}}, X_{\text{FC}}).
\end{equation*} Now, the goal becomes estimating/obtaining these latent variables 
$\boldsymbol{z}_{\text{joint}}, \boldsymbol{z}_{\text{sc}}^{\text{ind}},$ and $ \boldsymbol{z}_{\text{fc}}^{\text{ind}}$ from $X_{\text{SC}}, X_{\text{FC}}$. 

We design dual VAE encoders to achieve this goal. First, two modality-specific encoders independently produce parameters of diagonal Gaussian distributions: 
\begin{align*}
q_{\phi}(\tilde{\boldsymbol{z}}_{\text{fc}}^{\text{full}} \mid X_{\text{FC}}) &= \mathcal{N}\left(\boldsymbol{\mu}_{\text{fc}}, \text{diag}(\boldsymbol{\sigma}^2_{\text{fc}})\right), \\
q_{\phi}(\tilde{\boldsymbol{z}}_{\text{sc}}^{\text{full}} \mid X_{\text{SC}}) &= \mathcal{N}\left(\boldsymbol{\mu}_{\text{sc}}, \text{diag}(\boldsymbol{\sigma}^2_{\text{sc}})\right),
\end{align*}
Note that $\tilde{\boldsymbol{z}}_{\text{fc}}^{\text{full}}$ and $\tilde{\boldsymbol{z}}_{\text{sc}}^{\text{full}}$ capture modality-specific information independently and we have not separated the joint latent factor $\boldsymbol{z}_{\text{joint}}$, SC-specific factor $\boldsymbol{z}_{\text{sc}}^{\text{ind}}$, and FC-specific $\boldsymbol{z}_{\text{fc}}^{\text{ind}}$ yet. In other words, $\tilde{\boldsymbol{z}}_{\text{fc}}^{\text{full}} \neq [\boldsymbol{z}_{\text{joint}}; \boldsymbol{z}_{\text{fc}}^{\text{ind}}]$ and $\tilde{\boldsymbol{z}}_{\text{sc}}^{\text{full}} \neq [\boldsymbol{z}_{\text{joint}}, \boldsymbol{z}_{\text{sc}}^{\text{ind}}]$. The key challenge lies in transforming these unimodal representations into the disentangled components that satisfy the orthonormal constraints. The following subsection presents our attention-based fusion mechanism that achieves this disentanglement.

\subsection{Cross-Modal Joint-Individual Separation Network}\label{sec:CM-JIVE}

The encoder outputs $\tilde{\boldsymbol{z}}_{\mathrm{fc}}^{\mathrm{full}},\;
 \tilde{\boldsymbol{z}}_{\mathrm{sc}}^{\mathrm{full}}
 \in \mathbb{R}^{d_z}$
are produced independently without cross-modal information exchange. To decompose them into joint and individual components $(\boldsymbol{z}_{\mathrm{joint}}, \boldsymbol{z}_{\text{sc}}^{\mathrm{ind}}, \boldsymbol{z}_{\text{fc}}^{\mathrm{ind}})$, we must determine which features represent common patterns across both modalities versus which are modality-specific.

To achieve the disentangled factorization, we develop multi-head self-attention to learn subject-specific cross-modal alignments. For each subject, we compute attention weights quantifying how each feature in $\tilde{\boldsymbol{z}}_{\text{fc}}^{\mathrm{full}}$ relates to each feature in $\tilde{\boldsymbol{z}}_{\text{sc}}^{\mathrm{full}}$. Features exhibiting strong cross-modal correlation should contribute to the joint component $\boldsymbol{z}_{\mathrm{joint}}$, while weakly correlated features should remain in the individual components $\boldsymbol{z}_{\text{fc}}^{\mathrm{ind}}$ and $\boldsymbol{z}_{\text{sc}}^{\mathrm{ind}}$.

\subsubsection{Attention-based Fusion}

We first project the latent features into a shared space through modality-specific linear transformations
$
\boldsymbol{h}_{\text{fc}} = \mathbf{W}_{\text{FC}} \tilde{\boldsymbol{z}}_{\text{fc}}^{\mathrm{full}}\in \mathbb{R}^{d_z}, \quad
\boldsymbol{h}_{\text{sc}} = \mathbf{W}_{\text{SC}} \tilde{\boldsymbol{z}}_{\text{sc}}^{\mathrm{full}}
\in \mathbb{R}^{d_z}$ and then stack them into a matrix token
$
\mathbf{H} = [\boldsymbol{h}_{\text{fc}}, \boldsymbol{h}_{\text{sc}}]
$. This token will then be processed using multiple attention heads: $\mathbf{head}^{(j)} = \text{Attention}(Q^j, K^j, V^j)$, where $Q^j, K^j, V^j$ are queries, keys and values obtained via linear transformations of $\mathbf{H}$.
Outputs from all attention heads are concatenated and linearly transformed into $\widetilde{\mathbf{H}} = [\tilde{\boldsymbol{h}}_{\text{fc}}, \tilde{\boldsymbol{h}}_{\text{sc}}]$, where $\widetilde{\mathbf{H}} \in \mathbb{R}^{2 \times d_z}$.

\subsubsection{Joint and Individual Components Separation}

To allow flexible adjustment of each modality's contribution, we introduce a learnable gating
modulation to rescale each column in $\widetilde{\mathbf{H}}$ and then fuse them based on $
\boldsymbol{h}_{\mathrm{fused}}
=
\hat{\boldsymbol{h}}_{\text{fc}}
+
\hat{\boldsymbol{h}}_{\text{sc}},
$
where $\hat{\boldsymbol{h}}_j=m_j\tilde{\boldsymbol{h}}_j$ for $j\in\{\text{fc},\text{sc}\}$ and $m_j\in(0,1)$ is learned through a sigmoid gating operation from $\tilde{\boldsymbol{h}}_j$ .  
The fused representation $\boldsymbol{h}_{\mathrm{fused}} \in \mathbb{R}^{d_z}$ summarizes
the shared information from both modalities.

The fused vector is then passed through a linear transformation that expands its dimensionality
to $6d_z$, yielding
 $\mathbf{v} \in \mathbb{R}^{6d_z}$.
This vector is then partitioned into six contiguous segments of size $d_z$:
\[
\big(
\mathbf{v}_{1},\mathbf{v}_{2},\mathbf{v}_{3},
\mathbf{v}_{4},\mathbf{v}_{5},\mathbf{v}_{6}
\big)
=
\mathrm{Split}(\mathbf{v}),
\qquad
\mathbf{v}_{\ell}\in\mathbb{R}^{d_z}.
\]
Each sliced segment is subsequently passed through a modality-specific linear transformation
to obtain the final posterior parameters, $(\boldsymbol{\mu}_{\mathrm{joint}}, \log\boldsymbol{\sigma}_{\mathrm{joint}}, \boldsymbol{\mu}_{\text{fc}}^{\mathrm{ind}}, \log\boldsymbol{\sigma}_{\text{fc}}^{\mathrm{ind}},\boldsymbol{\mu}_{\text{sc}}^{\mathrm{ind}},\log\boldsymbol{\sigma}_{\text{sc}}^{\mathrm{ind}})$. These parameters can be used to generate joint and individual embeddings, $\boldsymbol{z}_{\mathrm{joint}},\boldsymbol{z}_{\text{fc}}^{\mathrm{ind}},$ and $\boldsymbol{\mu}_{\text{sc}}^{\mathrm{ind}}$.

To ensure that the three components capture non-overlapping information, we constrain them to be orthogonal to each other with the following loss function:
\begin{equation}
\begin{aligned}
\mathcal{L}_{\mathrm{ortho}}
&=
\cos^{2}\!\left(\boldsymbol{\mu}_{\mathrm{joint}}, \boldsymbol{\mu}_{\text{fc}}^{\mathrm{ind}}\right)
+
\cos^{2}\!\left(\boldsymbol{\mu}_{\mathrm{joint}}, \boldsymbol{\mu}_{\text{sc}}^{\mathrm{ind}}\right) \\
&\quad+
\cos^{2}\!\left(\boldsymbol{\mu}_{\text{fc}}^{\mathrm{ind}}, \boldsymbol{\mu}_{\text{sc}}^{\mathrm{ind}}\right),
\end{aligned}
\label{eq:orthogonality_loss}
\end{equation}
where 
$
\cos(\mathbf{a},\mathbf{b})
=
{\langle \mathbf{a}, \mathbf{b} \rangle}/(
{\|\mathbf{a}\|\,\|\mathbf{b}\| + \varepsilon})$ for a small $\varepsilon>0$. 

Unlike classical JIVE \cite{Lock2013} that relies on linear projections, multi-head attention captures complex, nonlinear relationships between modalities. Moreover, classical JIVE decomposes each data matrix into joint structure, individual structure, and noise via SVD-based constrained factorization requiring orthogonal subspaces. CM-JIVNet achieves an analogous decomposition $\boldsymbol{z} = [\boldsymbol{z}_{\text{joint}}; \boldsymbol{z}_{\text{fc}}^{\text{ind}}; \boldsymbol{z}_{\text{sc}}^{\text{ind}}]$ in latent space with three advantages: (i) nonlinearity via deep networks captures complex SC-FC relationships beyond linear correlation; (ii) probabilistic treatment quantifies decomposition uncertainty; (iii) attention-based learning enables subject-specific rather than population-average factorizations. The orthogonality constraint in Eq.~\eqref{eq:orthogonality_loss} mirrors JIVE's orthogonality requirement but operates in the latent space rather than observed data.

\subsection{Dual VAE Decoders}\label{subsec:decoders}

The disentangled latent components or scores must be decoded back to connectivity matrices to enable reconstruction-based training and the generative nature of the proposed model. We employ modality-specific decoders that map from latent scores to brain networks: the FC decoder reconstructs functional connectivity from $\boldsymbol{z}_{\text{fc}}^{\text{full}} = [\boldsymbol{z}_{\text{joint}}; \boldsymbol{z}_{\text{fc}}^{\text{ind}}]$, while the SC decoder reconstructs structural connectivity from $\boldsymbol{z}_{\text{sc}}^{\text{full}} = [\boldsymbol{z}_{\text{joint}}; \boldsymbol{z}_{\text{sc}}^{\text{ind}}]$. 

Assuming each brain network is formed by $V$ nodes (brain regions) and $D=V(V-1)/2$ number of unique connections (ignoring self-connections). For edge (or connection) $\ell$, let $X_{FC}[\ell]$ denote its FC connection strength, and $[u(\ell), v(\ell)]$ denote its node pair with $u(\ell) < v(\ell)$. We model each edge observation conditionally independent given the latent scores:
\begin{align*}
p_\theta(X_{\text{FC}} \mid \boldsymbol{z}_{\text{fc}}^{\text{full}}) &= \prod_{\ell=1}^{D} p_\theta(X_{\text{FC}}[\ell] \mid \boldsymbol{z}_{\text{fc}}^{\text{full}}) \\
p_\theta(X_{\text{SC}} \mid \boldsymbol{z}_{\text{sc}}^{\text{full}}) &= \prod_{\ell=1}^{D} p_\theta(X_{\text{SC}}[\ell] \mid \boldsymbol{z}_{\text{sc}}^{\text{full}}).
\end{align*}

\subsubsection{Functional Connectivity Decoder}
In the FC decoder, we model each connection with a Gaussian likelihood (note that we have applied Fisher $z$-transformation in the preprocessing step):
\begin{equation*}\label{eq:fc_likelihood}
p_\theta(X_{\text{FC}} \mid \boldsymbol{z}_{\text{fc}}^{\text{full}}) = \prod_{\ell=1}^D \mathcal{N}(X_{\text{FC}}[\ell]; \mu^F_{\ell}(\boldsymbol{z}_{\text{fc}}^{\text{full}}), \sigma_F^2),
\end{equation*}
where $\mu^F_{\ell}(\boldsymbol{z}_{\text{fc}}^{\text{full}})$ is the mean for $\ell$-th edge, and $\sigma_F^2$ is a learnable variance parameter for all edges. The edge mean $\mu^F_{\ell}$ is a function of $\boldsymbol{z}_{\text{fc}}^{\text{full}}$. We employ a two-stage process to compute it: first, map the latent vector to node-level representations, and then compute edge predictions from pairwise node interactions. Specifically, we transform $\boldsymbol{z}_{\text{fc}}^{\text{full}}$ into a node-level embedding ${X}^{F} \in \mathbb{R}^{V\times R_F}$  via a ResNet block.
The $r$-th column ${X}^{F}[\cdot,r] \in \mathbb{R}^{V}$ assigns an embedding value to each of the $V$ brain regions. 
The edge mean is then computed as a weighted sum of products between node-level embeddings:
\begin{equation*}\label{eq:fc_bilinear}
\mu^{F}_{\ell} = b^{F}_{\ell} + \sum_{r=1}^{R_F} \alpha^{F}_{r} \, {X}^{F}[u(\ell),r] \, {X}^{F}[v(\ell),r],
\end{equation*}
where $b^{F}_{\ell}$ is an edge-specific bias term and $\alpha^{F}_{r} \geq 0$ are learned weights.

With $\mu^{F}_{\ell}$ and $\sigma_F$, we can easily write down the log-likelihood for FC as:
\begin{equation} \label{eq:fc_ll}
\begin{aligned}
& \log p_{\theta}(X_{\text{FC}} \mid \boldsymbol{z}_{\text{fc}}^{\text{full}}) \\
= & -\frac{D}{2}\log(2\pi\sigma_F^2) - \frac{1}{2\sigma_F^2} \sum_{\ell=1}^{D} (X_{\text{FC}}[\ell] - \mu^F_{\ell})^2. 
\end{aligned}
\end{equation}

\subsubsection{Structural Connectivity Decoder} The SC data is very different from FC data. In our case, the SC matrix contains discrete, nonnegative streamline counts, and therefore, we use a Poisson distribution to characterize the uncertainty:
\begin{equation*}\label{eq:sc_likelihood}
p_\theta(X_{\text{SC}} \mid \boldsymbol{z}_{\text{sc}}^{\text{full}}) = \prod_{\ell=1}^D \text{Poisson}(X_{\text{SC}}[\ell]; \lambda^S_{\ell}(\boldsymbol{z}_{\text{sc}}^{\text{full}})), 
\end{equation*}
where  $\lambda^S_{\ell}(\boldsymbol{z}_{\text{sc}}^{\text{full}})$ is the rate parameter for the Poisson distribution. To model  $\lambda^S_{\ell}$, we employ a similar two-stage process to what we do for the FC data. We first transform $\boldsymbol{z}_{\text{sc}}^{\text{full}}$ into node-level embeddings ${X}^{S} \in \mathbb{R}^{V \times R_S}$ through multi-layer perceptrons. 

Each row of ${X}^{S}$ assigns an embedding vector to each of the $V$ brain regions. Next, $\lambda^S_{\ell}$ is computed from node embeddings with an exponential link to ensure positivity:
\begin{equation*}\label{eq:sc_rate}
\lambda^{S}_{\ell}(\boldsymbol{z}_{\text{sc}}^{\text{full}}) = \exp\big\{\gamma^{S}_{\ell} + \sum_{r=1}^{R_S} \alpha^{S}_{r} \, {X}^{S}[u(\ell),r] \, {X}^{S}[v(\ell),r]\big\},
\end{equation*}
where $\gamma^{S}_{\ell}$ is an edge-specific bias term and $\alpha^{S}_{r} \geq 0$ are learned weights. The exponential function guarantees that $\lambda^{S}_{\ell} > 0$, satisfying the Poisson distribution requirement. With $\lambda^S_{\ell}$, we can write the log-likelihood for SC as:
\begin{equation} \label{eq:sc_ll}
\begin{aligned}
&\log p_{\theta}(X_{\text{SC}} \mid \boldsymbol{z}_{\text{sc}}^{\text{full}}) \\
& = \sum_{\ell=1}^{D} \left(
X_{\text{SC}}[\ell] \log \lambda^{S}_{\ell}
- \lambda^{S}_{\ell}
- \log(X_{\text{SC}}[\ell]!)
\right).
\end{aligned}
\end{equation}

\subsection{Model Training}
The CM-JIVNet is trained by maximizing a variational lower bound on the joint log-likelihood, augmented with regularization terms that enforce disentanglement. The complete objective is discussed below, and it balances reconstruction accuracy, distributional regularization, cross-modal information sharing, and component orthogonality. 

\subsubsection{Evidence Lower Bound}

We use the evidence lower bound (ELBO) to approximate the intractable log marginal likelihood $\log p_{\theta}(X_{\text{FC}}, X_{\text{SC}})$ as follows: 
\begin{equation*}\label{eq:elbo}
\begin{split}
& \log p_{\theta}(X_{\text{FC}}, X_{\text{SC}}) \\
\geq &  \mathbb{E}_{q_{\phi}}\left[\log p_{\theta}(X_{\text{FC}} \mid \boldsymbol{z}_{\text{fc}}^{\text{full}}) + \log p_{\theta}(X_{\text{SC}} \mid \boldsymbol{z}_{\text{sc}}^{\text{full}})\right] \\
&\quad - D_{\text{KL}}(q_{\phi}(\boldsymbol{z} \mid X_{\text{FC}}, X_{\text{SC}}) \| p(\boldsymbol{z})),
\end{split}
\end{equation*}
where $p(\boldsymbol{z}) = \mathcal{N}(\mathbf{0}, \mathbf{I})$ is the standard Gaussian prior, $q_{\phi}(\boldsymbol{z} \mid X_{\text{FC}}, X_{\text{SC}})$ is the approximate posterior produced by the encoders and fusion mechanism. The first term encourages accurate reconstruction of both modalities, while the second term regularizes the posterior toward the prior.

Utilizing the derived conditional log-likelihoods in Equation.~\ref{eq:fc_ll} and \ref{eq:sc_ll}, we approximate the first data term as 
\begin{equation*}\label{eq:recon_combined}
\mathcal{L}_{\text{recon}} = \lambda_{\text{FC}} \mathcal{L}_{\text{FC}} + \lambda_{\text{SC}} \mathcal{L}_{\text{SC}},
\end{equation*}
where 
$\mathcal{L}_{\text{FC}} = - \mathbb{E}_{q_{\phi}}\left[ \sum_{\ell=1}^{D} (X_{\text{FC}}[\ell] - \mu^F_{\ell})^2 \right]$ and 
$\mathcal{L}_{\text{SC}} = -\mathbb{E}_{q_{\phi}}\left[\sum_{\ell=1}^{D} \left(
X_{SC}[\ell] \log \lambda^{S}_{\ell}
- \lambda^{S}_{\ell}
\right) \right] $, and $\lambda_{\text{FC}}, \lambda_{\text{SC}} > 0$ balance the relative importance of each modality.

The KL divergence term is decomposed across the model's two-stage encoding architecture. First, we regularize the initial encoder distributions:
\begin{align*} \label{eq:kl_enc}
\mathcal{L}_{\text{KL,enc}}
&= D_{\text{KL}}(q_{\phi}(\tilde{\boldsymbol{z}}_{\text{fc}}^{\text{full}} \mid X_{\text{FC}}) \| p(\tilde{\boldsymbol{z}}_{\text{fc}})) \nonumber \\
&\quad + D_{\text{KL}}(q_{\phi}(\tilde{\boldsymbol{z}}_{\text{sc}}^{\text{full}} \mid X_{\text{SC}}) \| p(\tilde{\boldsymbol{z}}_{\text{sc}})).
\end{align*}
Next, we also regularize the disentangled distributions produced by the fusion step:
\begin{equation*} \label{eq:kl_fusion}
\begin{split}
\mathcal{L}_{\text{KL,fusion}} &= D_{\text{KL}}(q_{\phi}(\boldsymbol{z}_{\text{joint}} \mid X_{\text{FC}}, X_{\text{SC}}) \| p(\boldsymbol{z}_{\text{joint}})) \\
&\quad + D_{\text{KL}}(q_{\phi}(\boldsymbol{z}_{\text{fc}}^{\text{ind}} \mid X_{\text{FC}}, X_{\text{SC}}) \| p(\boldsymbol{z}_{\text{fc}}^{\text{ind}})) \\
&\quad + D_{\text{KL}}(q_{\phi}(\boldsymbol{z}_{\text{sc}}^{\text{ind}} \mid X_{\text{FC}}, X_{\text{SC}}) \| p(\boldsymbol{z}_{\text{sc}}^{\text{ind}})).
\end{split}
\end{equation*}
Both terms have a closed-form formula since the involved distributions are Gaussians.

\subsubsection{Additional Regularizations}

Beyond the ELBO, we impose two additional constraints to achieve effective disentanglement in the latent space.

\paragraph{Mutual Information Regularization} To ensure the initial encoders capture joint cross-modal information before disentanglement, we maximize their mutual information using the Mutual Information Neural Estimator (MINE) \cite{Belghazi2018MINE}. MINE provides a variational lower bound on mutual information:

\begin{align*} \label{eq:mine}
I(\tilde{\boldsymbol{z}}_{\text{fc}}^{\text{full}}; \tilde{\boldsymbol{z}}_{\text{sc}}^{\text{full}})
&\geq \mathbb{E}_{p}\left[T_{\psi}(\tilde{\boldsymbol{z}}_{\text{fc}}^{\text{full}}, \tilde{\boldsymbol{z}}_{\text{sc}}^{\text{full}})\right] - \log\left(\mathbb{E}_{p_{\text{prod}}}\left[e^{T_{\psi}(\tilde{\boldsymbol{z}}_{\text{fc}}^{\text{full}}, \tilde{\boldsymbol{z}}_{\text{sc}}^{\text{full}})}\right]\right),
\end{align*}
where $T_{\psi}: \mathbb{R}^{d} \times \mathbb{R}^{d} \to \mathbb{R}$ is a neural network critic parameterized by $\psi$, $p$ denotes the joint distribution, and $p_{\text{prod}}$ denotes the product of marginals. The estimate $\hat{I}_{\psi}(\tilde{\boldsymbol{z}}_{\text{fc}}^{\text{full}}; \tilde{\boldsymbol{z}}_{\text{sc}}^{\text{full}})$ is obtained by maximizing this variational objective with respect to the neural critic $T_{\psi}$. We then define the loss term as:
\begin{equation*} \label{eq:mi_loss}
\mathcal{L}_{\text{MI}} = -\hat{I}_{\psi}(\tilde{\boldsymbol{z}}_{\text{fc}}^{\text{full}}; \tilde{\boldsymbol{z}}_{\text{sc}}^{\text{full}}),
\end{equation*}
This regularization term encourages the encoders to preserve cross-modal dependencies that will subsequently be allocated to the joint component.

\paragraph{Orthogonality Constraint} To enforce non-overlapping information in the modality-specific components, we minimize the orthogonality loss $ \mathcal{L}_{\text{ortho}}$ defined in Eq.~\eqref{eq:orthogonality_loss}. 

\subsubsection{Complete Objective}

The final training objective combines all terms:
\begin{equation}\label{eq:total_loss}
\mathcal{L} = \mathcal{L}_{\text{recon}} + \mathcal{L}_{\text{KL,enc}} + \lambda_{\text{fusion}} \mathcal{L}_{\text{KL,fusion}} + \lambda_{\text{MI}} \mathcal{L}_{\text{MI}} + \lambda_{\text{ortho}} \mathcal{L}_{\text{ortho}}
\end{equation}
where $\lambda_{\text{fusion}}, \lambda_{\text{MI}}, \lambda_{\text{ortho}} > 0$ control the relative importance of fusion regularization, cross-modal information sharing, and component disentanglement, respectively. This objective enables end-to-end training of CM-JIVNet while achieving probabilistic decomposition of brain connectivity into interpretable joint and individual components.

\subsubsection{Training Algorithm}
Algorithm~\ref{alg:cm_jivnet} summarizes the optimization procedure. We briefly describe the encoder and decoder architectures below.

\paragraph{CNN encoders for FC and SC}
We treat each connectome (FC or SC) as a single-channel 68-by-68 image and encode it with a modality-specific CNN.
The FC encoder and SC encoder share the same layer configuration but do not share parameters.
Each encoder contains three Conv--ReLU--MaxPool blocks: the three convolution layers output 32, 64, and 128 channels, and each 2-by-2 max-pooling step halves the spatial size, so 68 becomes 34, then 17, then 8.
The final 128-by-8-by-8 feature map is flattened into a 8192-dimensional vector and fed into two linear heads that output a 64-dimensional mean vector and a 64-dimensional log-variance vector.

\paragraph{Attention-based fusion and latent decomposition}
The two modality latents are first projected to a common hidden width (\texttt{hidden\_dim}) using modality-specific linear layers, and then stacked as a two-token sequence.
A 3-head self-attention layer exchanges information between the two tokens, after which a small gating layer (a linear layer to one scalar followed by a sigmoid) produces one weight per modality to form a weighted fused representation.
From this fused representation, three pairs of linear heads predict the mean and log-variance for one joint latent and two modality-specific latents.

\paragraph{Decoders}
For reconstruction, each decoder takes the concatenation of the joint latent and the corresponding modality-specific latent (two 64-dimensional vectors concatenated).

To reconstruct FC, we use a convolutional decoder.
A linear layer reshapes the concatenated latent into a 64-by-17-by-17 feature map; we use 17-by-17 so that two successive upsampling stages (each doubling the resolution) recover exactly 68-by-68.
We apply two residual blocks (each block has two 3-by-3 convolutions with batch normalization and ReLU plus a skip connection), followed by two transposed-convolution layers that upsample 17 to 34 and then 34 to 68 while reducing channels from 64 to 32 to 16.
A final 3-by-3 convolution maps 16 channels to one channel, and a \texttt{tanh} activation produces the reconstructed FC image.

To reconstruct SC, we use an MLP decoder.
The concatenated latent is passed through two fully connected layers of width \texttt{hidden\_dim} with ReLU activations, projected to a 68-by-68 output.
A \texttt{softplus} activation is applied at the output to enforce non-negativity, yielding the reconstructed SC image.

\begin{algorithm}[ht]
\caption{Training the CM-JIVNet Model}
\label{alg:cm_jivnet}
\begin{algorithmic}[1]
    \Require Data \(\{(X_{\text{FC}}^i, X_{\text{SC}}^i)\}_{i=1}^N\), batch size \(m\), learning rate \(\eta\), and hyperparameters \(\{\lambda_{\text{fusion}}, \lambda_{\text{MI}}, \lambda_{\text{ortho}}, \lambda_{\text{FC}}, \lambda_{\text{SC}}\}\).
    \State Randomly initialize model parameters \(\theta\) and \(\phi\).
    \For{\(epoch = 1\) to \(\text{max\_epochs}\)}
        \State Shuffle training data and split into batches of size \(m\).
        \For{each mini-batch \(\{(X_{\text{FC}}^i, X_{\text{SC}}^i)\}_{i=1}^m\)}
            \State Encode \(X_{\text{FC}}^i\) and \(X_{\text{SC}}^i\) to obtain initial latent variables \(\tilde{\boldsymbol{z}}_{\text{fc}}^i, \tilde{\boldsymbol{z}}_{\text{sc}}^i\) using the reparameterization trick.
            \State Apply CM-JIVE attention-based fusion to disentangle into \(\boldsymbol{z}_{\text{joint}}^i, \boldsymbol{z}_{\text{fc}}^{\text{ind},i}, \boldsymbol{z}_{\text{sc}}^{\text{ind},i}\).
            \State Decode using \([\boldsymbol{z}_{\text{joint}}^i; \boldsymbol{z}_{\text{fc}}^{\text{ind},i}]\) and \([\boldsymbol{z}_{\text{joint}}^i; \boldsymbol{z}_{\text{sc}}^{\text{ind},i}]\) to reconstruct \(X_{\text{FC}}^i\) and \(X_{\text{SC}}^i\).
            \State Compute \(\mathcal{L}_{\text{total}}\) using Eq.~\ref{eq:total_loss}.
            \State Update \(\theta\) and \(\phi\) via stochastic gradient-based optimization (Adam) with learning rate $\eta$.
        \EndFor
    \EndFor
    \State \Return Trained parameters \(\theta\) and \(\phi\).
\end{algorithmic}
\end{algorithm}

\subsection{Supervised CM-JIVNet}\label{subsec:supervised}
A central goal in network neuroscience is to decipher the relationship between individual brain connectivity and behavioral or cognitive phenotypes. For instance, to what extent can SC and FC patterns jointly predict an individual’s IQ? While the unsupervised CM-JIVNet encodes SC and FC into latent variables to capture intrinsic data variance, these representations may not fully encompass the specific features required for behavioral prediction. To address this, we extend CM-JIVNet into a supervised framework (sCM-JIVNet). This variant allows the encoders to adapt to the downstream prediction task, extracting latent variables that are not only statistically representative but also behaviorally relevant. In sCM-JIVNet, we leverage the pretrained encoder and fusion modules as a robust initialization and add linear regression heads to guide the latent space toward trait prediction.

For a given subject's data $X_{\text{FC}}^i$ and $X_{\text{SC}}^i$,  sCM-JIVNet utilizes the pretrained VAE encoders  in CM-JIVNet to obtain three disentangled latents:
$\boldsymbol{\mu}_{\text{joint}}^i$, $\boldsymbol{\mu}_{\text{fc}}^{\text{ind},i}$, and $\boldsymbol{\mu}_{\text{sc}}^{\text{ind},i}$. These latents are then concatenated into a vector $\boldsymbol{z}_{\text{cat}}^i$ for trait prediction. We use means rather than sampled latents to get deterministic predictions and avoid stochasticity during inference.

\subsubsection{Prediction Heads and Revised Training Objective}
For each target trait $t_k$ in the trait set $\mathcal{T} = \{t_1, \ldots, t_K\}$, we attach a linear regression head:
\begin{equation*}\label{eq:prediction}
y_k^i = \boldsymbol{\beta}_k^\top \boldsymbol{z}_{\text{cat}}^i + b_k + \epsilon_{k}^i,
\end{equation*}
where $\boldsymbol{\beta}_k$ and $b_k$ are learnable parameters, and $\epsilon_{k}^i$ captures the prediction error. We use a composite loss that balances mean squared error and correlation to train the prediction model:
\begin{equation*} \label{eq:sup_loss}
\mathcal{L}_{\text{sup}} = \sum_{k=1}^{K} \left( \lambda_{\text{mse}} \text{MSE}(\hat{y}_k, y_k) + \lambda_{\text{corr}} (1 - \text{corr}(\hat{y}_k, y_k)) \right).
\end{equation*}

\subsubsection{Two-Stage Fine-Tuning}

The sCM-JIVNet's training proceeds in two stages based on the pretrained CM-JIVNet:

\paragraph{Stage 1: Initialization} The pretrained encoders and fusion module are frozen, and only the regression heads $\{\boldsymbol{\beta}_k, b_k\}_{k=1}^K$ are optimized based on $\mathcal{L}_{\text{sup}}$. This initializes the prediction heads using pretrained CM-JIVNet.

\paragraph{Stage 2: Fine-Tuning.} All model parameters are then unfrozen and jointly optimized with:
\begin{equation*}\label{eq:joint_loss}
\mathcal{L}_{\text{s}} = \mathcal{L}_{\text{sup}} + \lambda \mathcal{L},
\end{equation*}
where $\lambda \ll 1$ applies weak reconstruction regularization to prevent catastrophic forgetting of the generative structure while allowing the encoders to adapt to the supervised task.

\section{Numerical Study}
We use the HCP-YA dataset introduced in Section \ref{sec:data} containing 1,065 subjects with paired FC and SC  to study the numerical performance of the proposed CM-JIVNet framework. With these data, we aim to address the following objectives: (1) validating that CM-JIVNet effectively disentangles joint and modality-specific latent components that capture interpretable brain organization patterns; (2) demonstrating that CM-JIVNet outperforms existing methods in cross-modality reconstruction while preserving both individual-level and group-level connectome patterns; and (3) evaluating whether the learned latent representations can predict behavioral traits and whether supervision improves predictive performance compared to unsupervised feature extraction.

\subsection{Validating Latent Space Disentanglement }
A key contribution of CM-JIVNet is its ability to decompose the latent representation into three components: a joint component capturing cross-modal information shared between FC and SC, and two individual components preserving modality-specific features unique to FC or SC. In this section, we first examine whether the proposed attention-based fusion successfully achieves this disentanglement.

To evaluate orthogonality, we quantify the geometric separation of the three latent subspaces by computing pairwise angles between latent vectors. For each subject $i$, CM-JIVNet produces three latent mean vectors: $\boldsymbol{\mu}^{i}_{\text{joint}} \in \mathbb{R}^{d_z}$, $\boldsymbol{\mu}^{\text{ind},i}_{\text{fc}} \in \mathbb{R}^{d_z}$, and $\boldsymbol{\mu}^{\text{ind},i}_{\text{sc}} \in \mathbb{R}^{d_z}$. For each pair of vectors, we compute their angle based on  
$\arccos\left( \frac{\langle \boldsymbol{\mu}_{a}, \boldsymbol{\mu}_{b} \rangle}{\|\boldsymbol{\mu}_{a}\| \|\boldsymbol{\mu}_{b}\|} \right)
$. We then aggregate these angles across all subjects, and report the mean and standard deviation. 

\begin{table}[h!]
\centering
\caption{Mean and standard deviation of pairwise angles between mean latent vectors (in degrees)}
\label{tab:latent_angles}
\begin{tabular}{lcc}
\toprule
Pair & Mean Angle (°) & Std. Dev. (°) \\
\midrule
$(\boldsymbol{\mu}_{\text{joint}},\boldsymbol{\mu}^{\text{ind}}_{\text{fc}})$ & 88.96 & 6.33 \\
$(\boldsymbol{\mu}_{\text{joint}},\boldsymbol{\mu}^{\text{ind}}_{\text{sc}})$ & 91.10 & 6.96 \\
$(\boldsymbol{\mu}^{\text{ind}}_{\text{fc}},\boldsymbol{\mu}^{\text{ind}}_{\text{sc}})$  & 89.20 & 6.96 \\
\bottomrule
\end{tabular}
\end{table}

Table~\ref{tab:latent_angles} shows that the mean angles are close to $90^\circ$ for all three pairs. These near-orthogonal angles confirm that the learned subspaces capture largely non-overlapping information. The relatively small standard deviations indicate that this orthogonality is consistent across subjects. This geometric separation validates that our attention-based separation mechanism, combined with the orthogonality loss, achieves effective disentanglement in the latent space.

\begin{figure}[h!]
  \centering
  \includegraphics[width=0.6\columnwidth]{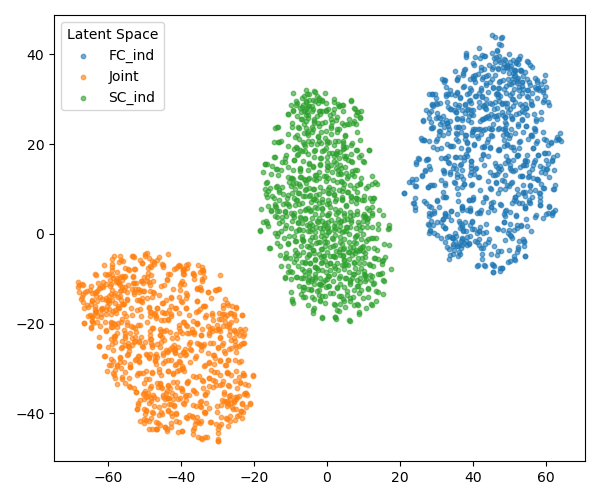}
  \caption{t-SNE embedding of concatenated latent representations, colored by subspace.}
  \label{fig:tsne_space}
\end{figure}

We next visualize the latent space separation using the t-SNE method. We vertically stack the three latent components from all subjects into a single matrix $\mathbf{X}_{\text{concat}} = [\boldsymbol{\mu}^{1:N}_{\text{joint}}; \boldsymbol{\mu}^{1:N,\text{ind}}_{\text{fc}}; \boldsymbol{\mu}^{1:N,\text{ind}}_{\text{sc}}] \in \mathbb{R}^{3N \times d_z}$. This row-wise concatenation allows t-SNE to embed every joint and modality-specific vector into a shared two-dimensional space. As shown in Fig.~\ref{fig:tsne_space}, the representations partition into distinct manifolds, confirming the model's ability to successfully disentangle cross-modal from modality-specific features.

Next, we quantify edge-wise SC--FC coupling induced by a single axis of the learned
joint latent space.
Let $\{\boldsymbol{\mu}^{i}_{\mathrm{joint}}\}_{i=1}^{N}$ denote the joint mean embeddings, and let
$\mathbf{u}_1$ be the first principal direction obtained by PCA.
We perform a conditional joint traversal by varying only the joint coordinate
$\boldsymbol{z}_{\mathrm{joint}}(t)
=
\bar{\boldsymbol{\mu}}_{\mathrm{joint}} + t\,\mathbf{u}_1$, while fixing the modality-specific latent variables.

\begin{figure}[t]
    \centering
    \includegraphics[width=0.7\columnwidth]{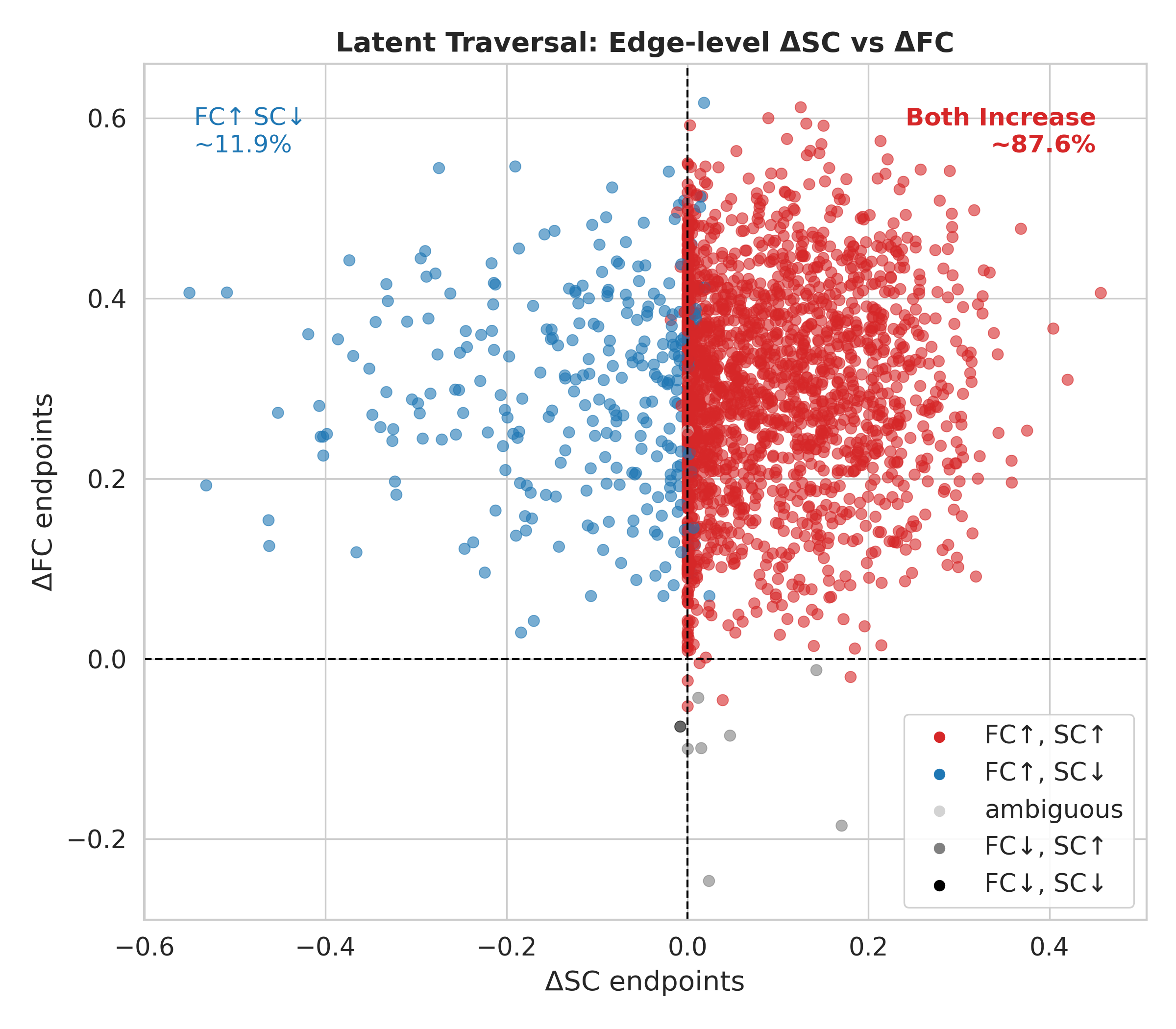}
    \caption{Edge-level SC-FC coupling induced by a PC$_1$ traversal in the joint latent space.
    Each point corresponds to one undirected edge.
    The horizontal and vertical axes report endpoint changes $\Delta\mathrm{SC}_{ij}$ and $\Delta\mathrm{FC}_{ij}$, respectively, between the traversal extremes (e.g., $\pm 3\sigma$ along PC$_1$).
    Quadrants categorize coupling signatures: $\mathrm{FC}\uparrow/\mathrm{SC}\uparrow$ (concordant increase), $\mathrm{FC}\uparrow/\mathrm{SC}\downarrow$ (decoupling), and the remaining sign combinations.
    A dominant mass in the $\mathrm{FC}\uparrow/\mathrm{SC}\uparrow$ quadrant indicates that the leading joint axis encodes a coupled mode of variation, while the minority decoupling edges highlight edge-specific departures from monotone SC--FC co-variation.}
    \label{fig:edge_coupling_scatter}
\end{figure}

Decoding yields $\mathrm{FC}(t)$ and $\mathrm{SC}(t)$ in the original network space.
For each edge $(i,j)$, we measure the endpoint changes
$\Delta \mathrm{FC}_{ij} \triangleq \mathrm{FC}_{ij}(t_{+})-\mathrm{FC}_{ij}(t_{-})$
and
$\Delta \mathrm{SC}_{ij} \triangleq
\log(1+\mathrm{SC}_{ij}(t_{+}))-\log(1+\mathrm{SC}_{ij}(t_{-}))$,
where $(t_{-},t_{+})$ correspond to the traversal endpoints where set to $\pm 3\sigma$ along PC$_1$.As shown in Fig.~\ref{fig:edge_coupling_scatter}, the vast majority of edges ($\approx 87.6\%$) fall into the
$\{\Delta \mathrm{FC}_{ij}>0,\Delta \mathrm{SC}_{ij}>0\}$ quadrant,
indicating that the dominant joint axis produces increased connections in both modalities.
A notable subset ($\approx 11.9\%$) exhibits $\Delta \mathrm{FC}_{ij}>0$ but $\Delta \mathrm{SC}_{ij}<0$,
revealing localized decoupling where  structural weakening accompanies functional strengthening.
The remaining quadrants contain negligible edges ($<0.6\%$).
This edge-level view complements matrix reconstructions by explicitly showing that coupling in the joint space is heterogeneous rather than uniform.

\begin{figure}[t]
    \centering
    \includegraphics[width=0.9\columnwidth]{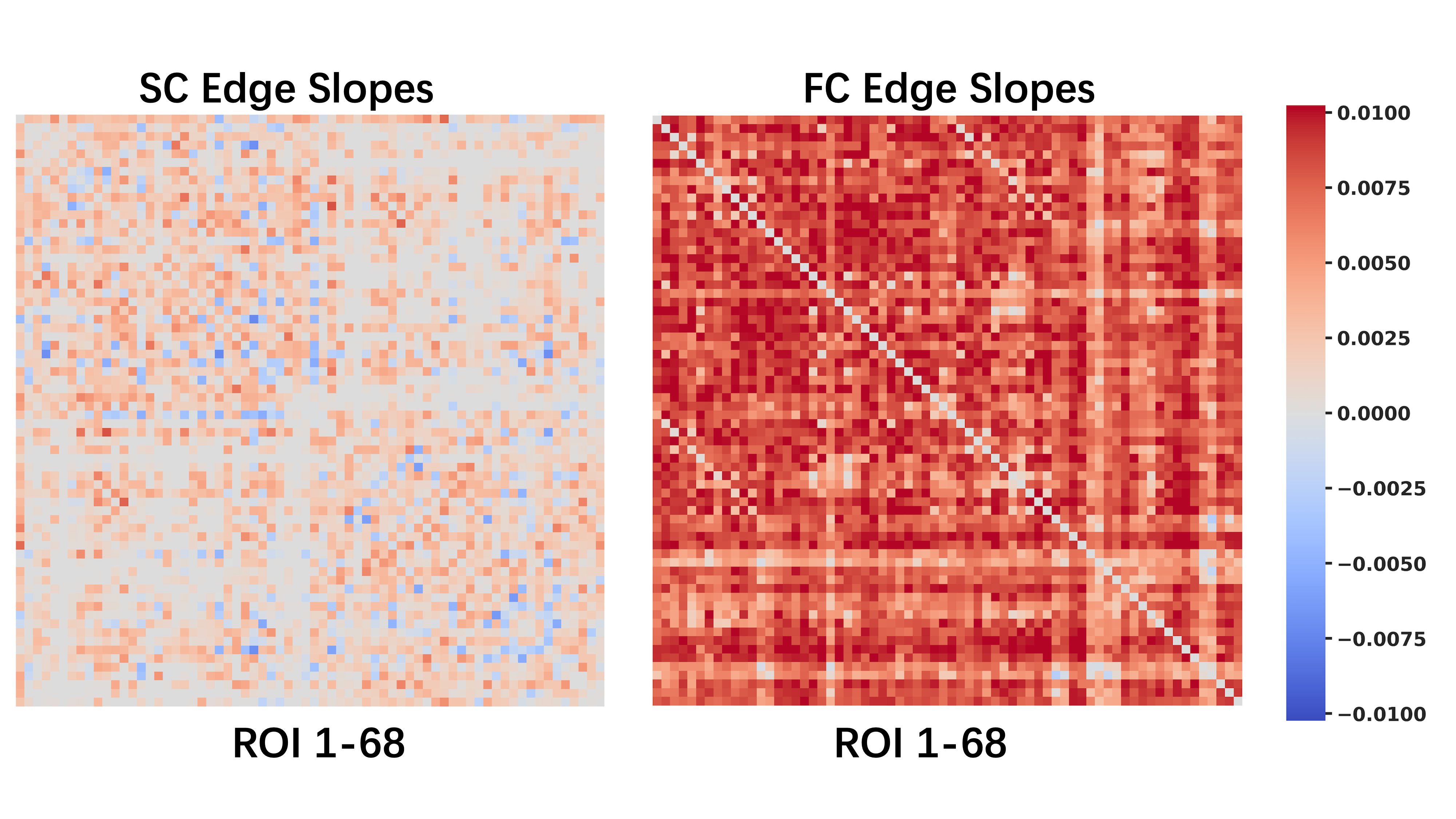}
    \caption{Spatial distribution of per-edge change rates along the joint latent PC$_1$.
    Each matrix entry is the least-squares slope with respect to the PC$_1$ coordinate,
    $b^{\mathrm{FC}}_{ij}$ for FC and $b^{\mathrm{SC}}_{ij}$ for $\log(1+\mathrm{SC})$.Positive values (red) indicate edges that strengthen as PC$_1$ increases, while negative values (blue) indicate weakening
    }
    \label{fig:edge_slope_mats}
\end{figure}

To localize where the joint axis acts most strongly, we estimate per-edge
rates of change along PC$_1$.
For each edge, we fit a least-squares line across traversal steps,
$\mathrm{FC}_{ij}(t)\approx a^{\mathrm{FC}}_{ij}+b^{\mathrm{FC}}_{ij}t$
and
$\log(1+\mathrm{SC}_{ij}(t))\approx a^{\mathrm{SC}}_{ij}+b^{\mathrm{SC}}_{ij}t$,
and visualize the slope matrices $\{b^{\mathrm{FC}}_{ij}\}$ and
$\{b^{\mathrm{SC}}_{ij}\}$ in Fig.~\ref{fig:edge_slope_mats}.
The FC slope map is broadly positive, consistent with a global functional modulation captured
by PC$_1$, whereas SC slopes are weaker and spatially heterogeneous, suggesting that
structural variations aligned with the same joint factor are more localized.

\subsection{Reconstruction Performance Evaluation}
We next evaluate whether CM-JIVNet can accurately reconstruct FC and SC matrices through its encoder and decoder and whether it outperforms baseline methods. We compare CM-JIVNet against several baselines: CNN--MLP, which uses convolutional encoders with multi-layer perceptron decoders; MLP--Attention, which incorporates attention mechanisms into the MLP architecture; and Graph Auto-Encoding (GATE)~\cite{Liu2021}, a graph-based variational autoencoder designed for brain network analysis. We assess reconstruction quality using Mean Squared Error (\textbf{MSE}), Structural Similarity Index Measure (\textbf{SSIM}) \cite{Wang2004SSIM}, Fréchet Inception Distance (\textbf{FID}), and Pearson \textbf{correlation}.
MSE evaluates element-wise reconstruction accuracy, while SSIM measures structural similarity by accounting for global intensity, contrast, and spatial organization of connectivity patterns.
FID compares the distribution of reconstructed and empirical connectomes in a feature space rather than at the matrix level; we consider two variants: Spectral FID, computed from the leading eigenvalues of the normalized graph Laplacian to capture global topology, and Graph FID, computed from summary graph-theoretic statistics to reflect local and meso-scale organization.
Finally, Pearson correlation quantifies similarity of connectivity patterns at both the individual-subject and group-average levels.

\textbf{Both modalities are provided at inference.}
Table~\ref{tab:recon_all} reports reconstruction performance when both FC and SC are available at inference.
In this joint reconstruction setting, the encoded representations
$\boldsymbol{\mu}_{\text{joint}}$, $\boldsymbol{\mu}^{\text{ind}}_{\text{fc}}$, and
$\boldsymbol{\mu}^{\text{ind}}_{\text{sc}}$ are jointly used to reconstruct both modalities. For FC reconstruction, CM-JIVNet achieves the lowest MSE and the highest SSIM among all in-house baselines,
indicating improved edge-wise accuracy and structural similarity.
CM-JIVNet also yields the highest individual-level Pearson correlation
($0.816\pm0.046$), demonstrating superior preservation of subject-specific functional connectivity patterns.
While MLP--Attention attains a lower FC Spectral FID, CM-JIVNet substantially improves spectral alignment
relative to GATE (3.42 vs.\ 74.87), reflecting markedly reduced distributional mismatch in the eigenspectrum. For SC reconstruction, CM-JIVNet achieves the lowest MSE, the highest SSIM, and the lowest Graph FID, indicating more faithful recovery of both edge-wise values and topological properties of structural connectivity.
Although MLP--Attention attains slightly lower spectral discrepancy, CM-JIVNet consistently achieves the highest individual-level Pearson correlation ($0.923\pm0.024$), suggesting superior preservation of subject-specific structural organization beyond global spectral similarity.

\begin{table*}[!t]
\centering
\caption{In-house baseline comparison for FC and SC reconstruction (both modalities available at inference). All models use the same data split and preprocessing.}
\label{tab:recon_all}

\setlength{\tabcolsep}{3.2pt}
\renewcommand{\arraystretch}{1.15}
\footnotesize
\begin{tabular}{lcccccc ccccc}
\toprule
& \multicolumn{6}{c}{\textbf{FC reconstruction}} & \multicolumn{5}{c}{\textbf{SC reconstruction}} \\
\cmidrule(lr){2-7}\cmidrule(lr){8-12}
\textbf{Model} &
\textbf{MSE}$\downarrow$ &
\textbf{SSIM}$\uparrow$ &
\textbf{Spectral FID}$\downarrow$ &
\textbf{Individual}$\uparrow$ &
\textbf{Group}$\uparrow$ &
\textbf{} &
\textbf{MSE}$\downarrow$ &
\textbf{SSIM}$\uparrow$ &
\textbf{Graph FID}$\downarrow$ &
\textbf{Individual}$\uparrow$ &
\textbf{Group}$\uparrow$ \\
\midrule
CM-JIVNet
& \textbf{0.011}
& \textbf{0.761}
& 3.420
& \textbf{0.816$\pm$0.046}
& 0.993
&  &
\textbf{0.575}
& \textbf{0.828}
& \textbf{0.226}
& \textbf{0.923$\pm$0.024}
& 0.999 \\

CNN--MLP
& 0.022
& 0.603
& 8.790
& 0.666$\pm$0.066
& \textbf{0.994}
&  &
0.587
& 0.793
& 0.239
& 0.918$\pm$0.023
& 0.999 \\

MLP--Attention
& 0.018
& 0.644
& \textbf{2.580}
& 0.709$\pm$0.066
& 0.984
&  &
0.654
& 0.817
& 0.232
& 0.917$\pm$0.024
& 0.999 \\

GATE 
& 0.037
& 0.262
& 74.870
& 0.714$\pm$0.161
& 0.818
&  &
1.576
& 0.594
& 6.777
& 0.857$\pm$0.013
& 0.897 \\

\bottomrule
\end{tabular}
\end{table*}

\textbf{Only one modality is provided at inference.} In clinical settings where only one modality is available, CM-JIVNet enables the prediction of the missing modality by leveraging the joint latent component. We evaluated this capacity using edge-wise Pearson correlation between empirical and predicted FC at both the group and individual levels (Table~\ref{tab:sc2fc_literature}). 
CM-JIVNet achieves a group-level correlation of 0.990 and an individual-level correlation of $0.712 \pm 0.074$. To ensure a fair comparison, our results are grouped by parcellation scale ($V$), as the number of nodes significantly impacts the prediction and evaluation tasks. Within the $V \approx 68$ group, our model outperforms recent benchmarks, including Neudorf et al. \cite{Neudorf2022GraphNets} (0.69) and Yang et al. \cite{Yang2022} (0.572). Furthermore, CM-JIVNet remains competitive with high-resolution models (e.g., Chen et al. \cite{Chen2023GCN}, $V=400$). 

The high group correlation signifies the successful recovery of dominant group-level connectivity, while the individual-level performance highlights the model’s modality-transforming ability---specifically, its capacity to extract subject-specific functional information from structural data. These findings confirm that CM-JIVNet accurately models the structural-functional interrelationship within the joint latent space, enabling the effective reconstruction of missing modalities from the available one.

\begin{table*}[!t]
\centering
\caption{SC$\rightarrow$FC prediction performance (Pearson correlation) compared with prior studies, grouped by parcellation scale ($V$).}
\label{tab:sc2fc_literature}
\setlength{\tabcolsep}{6pt}
\renewcommand{\arraystretch}{1.2}
\footnotesize
\begin{threeparttable}
\begin{tabular}{lrrlcc}
\toprule
\textbf{Study} & \textbf{N} & \textbf{Nodes ($V$)} & \textbf{Approach} & \textbf{Group} & \textbf{Individual} \\
\midrule
\multicolumn{6}{l}{\textit{Standard Resolution ($V \approx 68$)}} \\
\textbf{Ours (CM-JIVNet)} & \textbf{1,058} & \textbf{68} & \textbf{CNN + attention fusion + MINE} & \textbf{0.990} & \textbf{0.712 $\pm$ 0.074} \\
Li et al.~\cite{Li2019PredictingFC}        & 1,058 & 68  & Graph convolutional net & $\approx$0.75$^\dagger$ & -- \\
Neudorf et al.~\cite{Neudorf2022GraphNets}   & 998 & 66 & Graph Nets & 0.94 & 0.69 \\
Yang et al.~\cite{Yang2022}            & 103 & 68 & Graph autoencoder & 0.96 & 0.572 \\
Sarwar et al.~\cite{Sarwar2021} & 103 & 68 & Multilayer perceptron & 0.9 $\pm$ 0.1 & 0.5 $\pm$ 0.1 \\

\midrule
\multicolumn{6}{l}{\textit{Medium Resolution ($V \approx 90$--100)}} \\
Hong et al.~\cite{Hong2023GCN}                & 360 & 90 & Graph convolutional net & -- & 0.716--0.724 \\
Benkarim et al.~\cite{Benkarim2022Riemannian} & 326 & 100 & Riemannian optimization & -- & 0.804 $\pm$ 0.060 \\

\midrule
\multicolumn{6}{l}{\textit{High Resolution ($V = 400$)}} \\
Chen et al.~\cite{Chen2023GCN}                & 404 & 400 & Graph convolutional net & 0.953 & 0.715 $\pm$ 0.052 \\

\bottomrule
\end{tabular}
\begin{tablenotes}[flushleft]
\footnotesize
\item $^\dagger$ Values are reported/approximated as in the referenced survey table.
\end{tablenotes}
\end{threeparttable}
\end{table*}

\subsection{Behavioral Trait Prediction}

Linking brain connectivity to behavioral traits is one of the most important tasks in neuroscience. While the standard CM-JIVNet is excellent at reconstructing brain networks, its latent variables are not automatically optimized to predict behavior. To address this, we use the supervised version, sCM-JIVNet, which fine-tunes the model to extract features that are specifically relevant to cognitive traits. In this section, we evaluate how well these supervised representations predict various behavioral phenotypes compared to standard unsupervised features.

 We selected seven traits spanning distinct behavioral domains to provide a comprehensive evaluation: fluid intelligence (FluidInt\_CR, FluidInt\_SI), language abilities (OralReading, PictureVocab), visuospatial reasoning (Visuospatial), and affective/psychiatric measures (AngerAggr, ChildConduct). For each continuous trait, we performed 5-fold cross-validated Ridge regression, with the regularization parameter selected via nested cross-validation within each training fold. Predictive performance is evaluated using the Pearson correlation coefficient between observed and out-of-fold predicted trait values:
$ r = \text{corr}(\mathbf{y}, \widehat{\mathbf{y}})$, 
where $\mathbf{y}$ and $\widehat{\mathbf{y}}$ are vectors of observed and predicted trait scores across all subjects. 

\begin{figure}[htbp]
  \centering
  \includegraphics[width=0.7 \columnwidth]{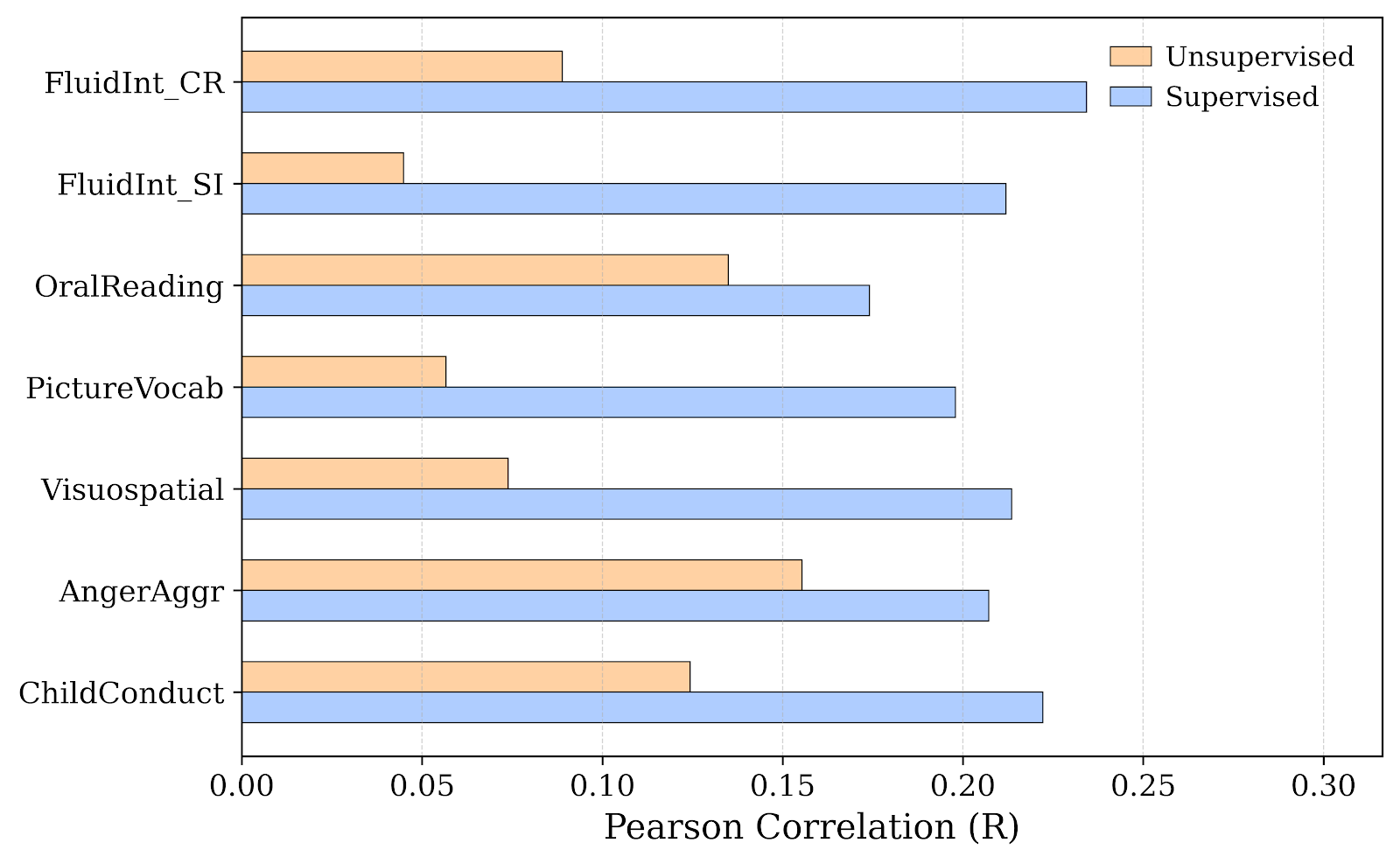}
  \caption{Comparison of behavioral trait prediction performances between unsupervised and supervised CM-JIVNet models using concatenated latent representations.}
  \label{fig:sup_unsup_concate}
\end{figure}

{Fig.~\ref{fig:sup_unsup_concate}} compares the prediction performance of the unsupervised and supervised CM-JIVNet models using the concatenated latent embeddings $\mathbf{z}^{i}_{\text{cat}}$. While the unsupervised model successfully captures intrinsic variance for connectome reconstruction, the {sCM-JIVNet} variant markedly improves behavioral prediction, achieving an average gain of $\Delta r \approx 0.11$. The most significant improvements are observed in Fluid Intelligence (FluidInt\_SI: $+0.17$, FluidInt\_CR: $+0.15$) and Language/Vocabulary (PictureVocab: $+0.14$). These results suggest that the latent features required to maximize reconstruction fidelity do not perfectly overlap with those that drive behavioral phenotypes. This finding aligns with recent literature suggesting that ``prediction-relevant'' brain features are often distinct from ``reconstruction-relevant'' ones, necessitating supervised objectives to bridge the gap between generative modeling and predictive neuroscience \cite{Liu2021, Zhang2024MotionInvariant}. By incorporating a supervised loss, the model effectively shifts its latent focus toward connectivity patterns that are biologically relevant to behavioral outcomes.

\begin{figure}[htbp]
  \centering
  \includegraphics[width=0.8 \columnwidth]{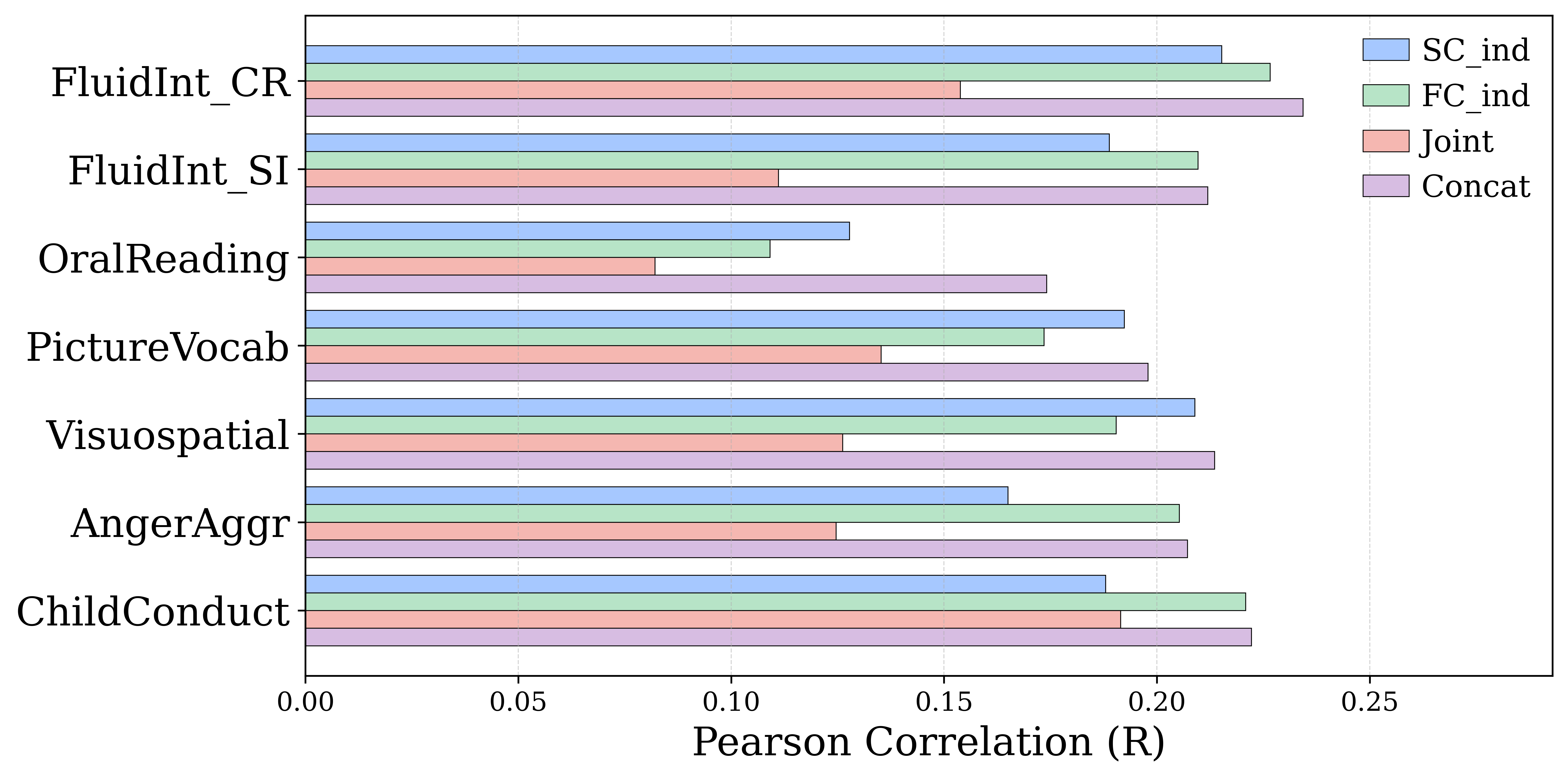}
  \caption{Comparison of behavioral trait prediction performance across latent feature configurations. Performance is reported as the Pearson correlation for joint, functional-individual, structural-individual, and concatenated subspaces.}
  \label{fig:latent_sup}
\end{figure}

To assess the specific contribution of each latent subspace, we compared four feature configurations: (1) joint latent variables ($\boldsymbol{\mu}_{\text{joint}}$), (2) FC-individual ($\boldsymbol{\mu}^{\text{ind}}_{\text{fc}}$), (3) SC-individual ($\boldsymbol{\mu}^{\text{ind}}_{\text{sc}}$), and (4) the concatenated set. As shown in {Figure~\ref{fig:latent_sup}}, the concatenated representation yields the highest accuracy across all traits, confirming that joint and modality-specific components provide complementary information. Specifically, {FC-individual} features demonstrate superior predictive power for fluid intelligence and affective traits, while {SC-individual} features perform on par with or slightly better than FC in language and visuospatial domains. This is consistent with evidence that functional co-activation patterns are more closely linked to dynamic cognitive processes like fluid reasoning \cite{dhamala2021distinct}, whereas structural white-matter integrity provides a more stable foundation for ``crystallized'' abilities like vocabulary and language syntax \cite{sanchez2023white}. Furthermore, the joint component captures shared SC-FC dependencies, often referred to as ``SC-FC coupling'', presents a fair prediction power with behavior traits, and has been identified as a critical neurobiological marker for general cognitive ability \cite{fotiadis2024structure}. These results validate that CM-JIVNet's disentangled latent space successfully isolates distinct, biologically meaningful aspects of brain organization.

\section{Discussion and Conclusion}
We have presented {CM-JIVNet}, a novel variational framework that integrates connectivity-tailored CNN-ResNet architectures with a JIVE-inspired joint--individual decomposition and multi-head attention fusion. By jointly modeling functional and structural connectomes on 1,065 HCP-YA subjects, our framework addresses a critical challenge in multimodal integration: the ability to disentangle shared cross-modal patterns from modality-specific variations within a non-linear, generative setting.

Our experimental results demonstrate that CM-JIVNet significantly outperforms standard generative baselines, including {GATE} and MLP-based architectures, across all reconstruction metrics. As shown in {Table II}, the model achieves the lowest MSE and highest SSIM, alongside substantially reduced spectral and graph FID scores, indicating superior preservation of both global topology and local graph properties. Furthermore, in the missing-modality scenario ({Table III}), CM-JIVNet achieves a state-of-the-art SC$\rightarrow$FC group-level correlation of {0.990} and an individual-level correlation of {0.712}, surpassing several recent graph-based and deep learning approaches in the literature.

A key contribution of this work is the {explicit disentanglement} of the latent space. Geometric analysis ({Table I}) and t-SNE embeddings ({Fig. 2}) verify that the joint and modality-specific subspaces partition into distinct, near-orthogonal manifolds. This separation is functionally meaningful: while the joint subspace captures fundamental organizational principles shared across modalities, the individual subspaces retain fine-grained details unique to structure or function. Latent traversal experiments (Figs.~\ref{fig:edge_coupling_scatter} and \ref{fig:edge_slope_mats}) further characterize this interpretability, revealing that the joint axis encodes a dominant mode of coordinated network strengthening, while simultaneously capturing spatially localized structural variations and specific SC--FC decoupling.

Additionally, our supervised extension, {sCM-JIVNet}, demonstrates that while unsupervised representations capture intrinsic data variance, targeted fine-tuning is essential for behavioral prediction. By incorporating a supervised objective, we achieved significant gains in predicting cognitive phenotypes, with an average improvement of $\Delta r \approx 0.11$. The model was particularly successful in identifying connectivity patterns relevant to fluid intelligence and language abilities ({Figs. 5--6}), validating that the disentangled components capture complementary aspects of brain organization relevant to human behavior.

Despite these successes, our study has certain limitations. We evaluated the framework on two modalities within a healthy young adult cohort; future research should test its utility in clinical populations and incorporate additional modalities, such as EEG or PET. While our VAE-based approach is highly effective, the integration of emerging generative architectures, such as {diffusion models}, may further enhance high-fidelity matrix synthesis. Finally, scaling CM-JIVNet to massive, diverse datasets like the {UK Biobank} and {ABCD study} will be critical to assess its generalizability across the human lifespan and in various disease states.

In conclusion, {CM-JIVNet} provides a robust, interpretable, and scalable solution for multimodal connectome analysis. By bridging the gap between deep generative modeling and factorized latent representations, this framework offers a powerful tool for uncovering the neural basis of cognition and developing clinically relevant biomarkers for brain disorders.

\bibliographystyle{IEEEtran}
\bibliography{myref}

\end{document}